\documentclass[sigconf,nonacm]{acmart}

\setcopyright{none}
\usepackage{booktabs}
\usepackage{makecell}
\usepackage{multirow}
\usepackage{xspace}
\usepackage{graphicx}
\usepackage{afterpage}
\usepackage[most]{tcolorbox}
\definecolor{okgreen}{HTML}{2E7D32}
\definecolor{badred}{HTML}{C0392B}
\newcommand{\cmk}{\textcolor{okgreen}{\checkmark}}
\newcommand{\xmk}{\textcolor{badred}{$\times$}}
\newcommand{\okhl}[1]{\textcolor{okgreen}{#1}}
\newcommand{\badhl}[1]{\textcolor{badred}{#1}}
\newtcolorbox{promptbox}[1]{enhanced, breakable,
  colback=black!3, colframe=black!55, boxrule=0.5pt, arc=1pt,
  left=5pt, right=5pt, top=2.5pt, bottom=2.5pt,
  title={#1}, fonttitle=\small\bfseries,
  fontupper=\scriptsize\ttfamily\raggedright,
  colbacktitle=black!55, coltitle=white}

\newcommand{\PNinetyFiveLatency}{10.73}

\newcommand{\MainbBcMgzJ}{61.01}
\newcommand{\MgScaleCmsJ}{58.70}
\newcommand{\MainbCatFiveNoRef}{444}
\newcommand{\MainbCatFiveAns}{2}

\title[EdgeMem]{EdgeMem: LLM-Free Agent Memory Construction and Retrieval via Evidence-Preserving Multi-Anchor Hypergraph}

\author{Zeyang Cui}
\email{ze-yang.cui@connect.polyu.hk}
\affiliation{%
  \institution{The Hong Kong Polytechnic University}
  \city{Hong Kong}
  \country{Hong Kong SAR}
}

\author{Jiannong Cao}
\email{jiannong.cao@polyu.edu.hk}
\affiliation{%
  \institution{The Hong Kong Polytechnic University}
  \city{Hong Kong}
  \country{Hong Kong SAR}
}

\author{Zhiyuan Wen}
\email{zyuanwen@polyu.edu.hk}
\affiliation{%
  \institution{The Hong Kong Polytechnic University}
  \city{Hong Kong}
  \country{Hong Kong SAR}
}

\author{Bo Yuan}
\authornote{Bo Yuan and Shengyuan Chen are co-corresponding authors.}
\email{yuanboit@chinamobile.com}
\affiliation{%
  \institution{JIUTIAN Research, China Mobile}
  \city{Beijing}
  \country{China}
}

\author{Junlan Feng}
\email{fengjunlanit@chinamobile.com}
\affiliation{%
  \institution{JIUTIAN Research, China Mobile}
  \city{Beijing}
  \country{China}
}

\author{Shengyuan Chen}
\authornotemark[1]
\email{sheng-yuan.chen@polyu.edu.hk}
\affiliation{%
  \institution{The Hong Kong Polytechnic University}
  \city{Hong Kong}
  \country{Hong Kong SAR}
}

\begin{document}

\begin{abstract}
Agent memory allows LLM agents to use earlier interactions when answering new
queries. Existing methods often compress interaction histories into summaries
or other LLM-generated representations. Repeated generation adds cost and can
discard answer-bearing details before the system knows what a future query will
require. We propose \textsc{EdgeMem}, an agent-memory method built around a
simple principle: preserve original interaction turns and organize them through
complementary content, temporal, and episodic cues. \textsc{EdgeMem} realizes
this principle with a multi-anchor hypergraph constructed by lightweight local
processing. Retrieval directly returns source evidence and reserves LLM use for
final answer generation, combining structured access to multi-session histories
with faithful retention of the original conversation. Experiments on LoCoMo
and LongMemEval-S show strong retrieval and memory-grounded question answering;
on LoCoMo, \textsc{EdgeMem} achieves the highest strict-judge score among seven
reproduced systems under a shared prompt (\MainbBcMgzJ{} versus
\MgScaleCmsJ{}), while construction and retrieval require no generative-LLM
calls. Overall, \textsc{EdgeMem} shows that preserving and organizing source
evidence provides an effective and efficient foundation for agent memory
without generative memory management.

\end{abstract}



\maketitle

\begin{figure}[t]
\centering
\includegraphics[width=\columnwidth]{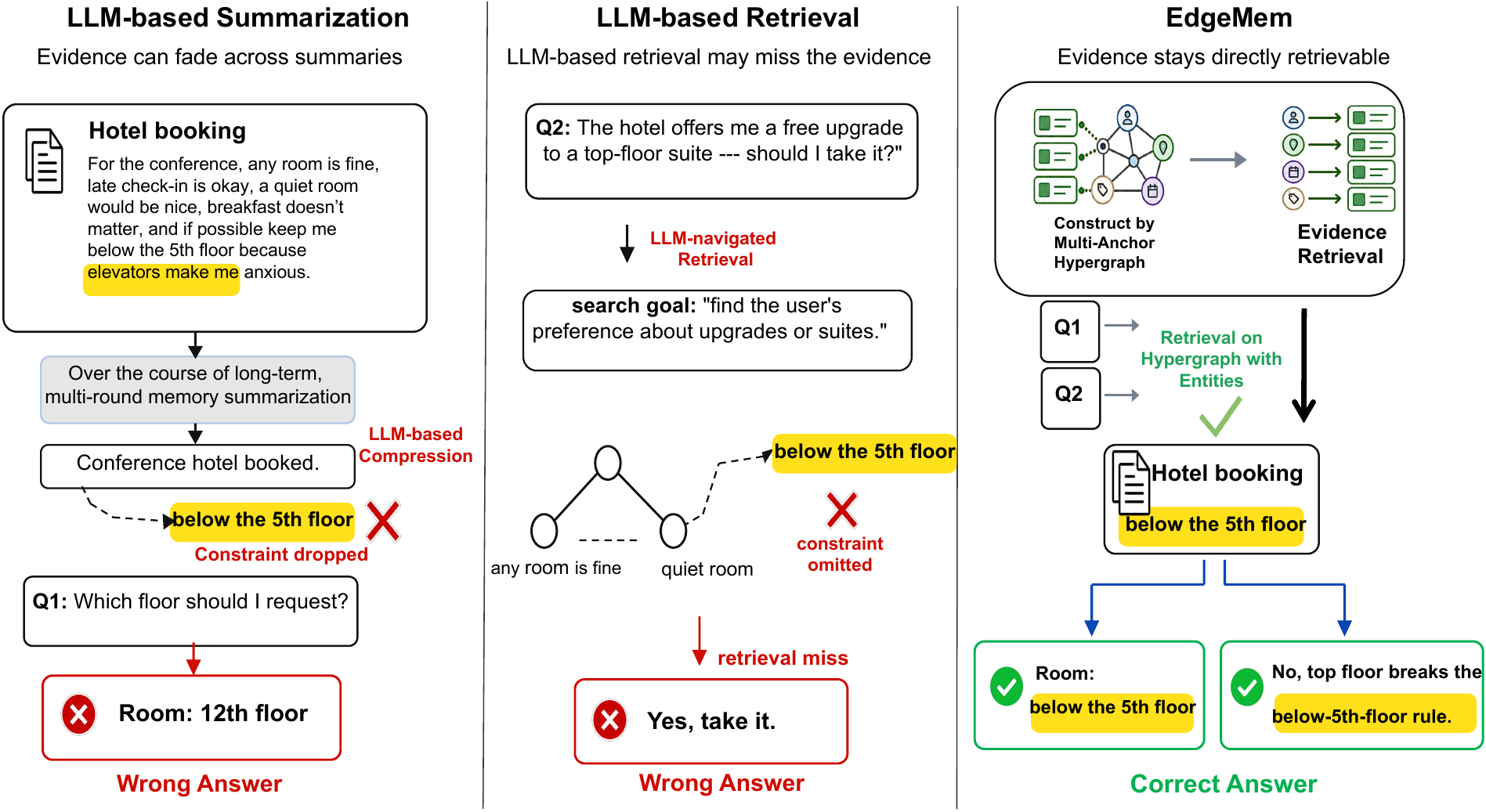}
\centerline{\small (a)}
\vspace{1pt}
\includegraphics[width=\columnwidth]{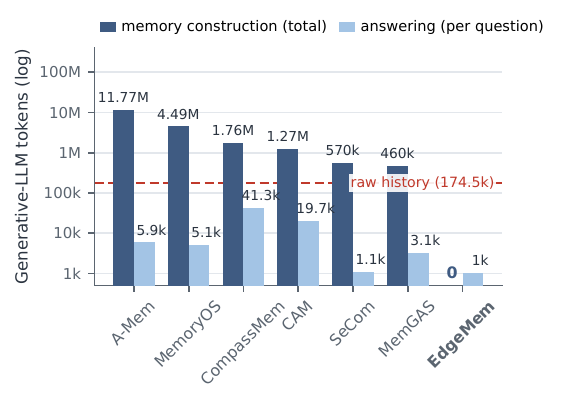}
\centerline{\small (b)}
\caption{\textbf{(a)}~LLM-based summarization and retrieval can lose
answer-bearing evidence, while original-evidence preservation keeps it
directly retrievable. \textbf{(b)}~Existing memory frameworks incur substantial generative-LLM
token costs in both memory construction and per-query answering.}
\vspace{-12pt}
\Description{Top: a side-by-side comparison in which LLM-based
summarization and retrieval drop an answer-bearing detail before answering,
while original-evidence preservation retrieves it directly. Bottom: a
grouped horizontal bar chart on a logarithmic token axis; EdgeMem's
construction cost is zero and its answering cost is the lowest.}
\label{fig:tokenstages}
\end{figure}

\section{Introduction}
\label{sec:intro}

Agent memory allows an LLM agent to draw on past interactions when answering a
new query \citep{zhang2024survey}. Across multi-session conversations,
relevant information may lie in a much earlier session
\citep{maharana2024locomo,wu2025longmemeval}. As the interaction history grows,
maintaining a usable memory and retrieving evidence relevant to the current
query become increasingly challenging. Feeding the full history to the model
for every query requires progressively more input tokens and may eventually
exceed the model's context window \citep{packer2023memgpt}. Agent memory systems
therefore commonly store past interactions externally and retrieve only a
small amount of relevant evidence for response generation
\citep{lewis2020rag,zhang2024survey}.

Existing agent-memory methods organize external histories through chunk
retrieval or derived memory representations. Conventional approaches segment
past interactions and retrieve relevant chunks using lexical or dense
similarity. More recent methods use LLM inference to extract or compress
salient information~\citep{chhikara2025mem0,pan2025secom}, construct
hierarchical stores, linked memory networks, and knowledge
graphs~\citep{xu2025amem,kang2025memoryos,gutierrez2025hipporag2}, or build
multi-granularity memory views~\citep{xu2026memgas}. LLMs may also guide
hierarchical navigation, plan retrieval, or filter candidate evidence at query
time~\citep{li2025cam,hu2026compassmem,xu2026memgas}. Methods that place LLM
inference inside the memory lifecycle introduce two limitations, as summarized in
Figure~\ref{fig:tokenstages}. First, repeated LLM calls for construction,
updating, and retrieval incur substantial token cost as the history grows. At
the evaluated LoCoMo scale, memory construction alone already consumes more
tokens than the raw history for every mainstream baseline
(Figure~\ref{fig:tokenstages}b). Second, generated summaries and extracted
memory units may omit or distort details from the original turns
\citep{pan2025secom,zhuang2025linearrag}. Once an answer-bearing detail is lost,
later retrieval stages cannot recover it
(Figure~\ref{fig:tokenstages}a). Together, these limitations motivate a memory
design that organizes original conversational evidence directly and reserves
generation for the final answer.

Inspired by episodic-memory research, which characterizes an occurrence through
its content and spatiotemporal context
\citep{tulving1972episodic,clayton1998episodic}, we view conversational memory
as original interaction records organized by complementary access cues. A turn
can be recalled through what it discusses, when it occurs, and the episode in
which it appears. We therefore retain each turn as an atomic evidence record
and associate it with multiple anchors for content, time, and episode context.
A query can then follow the anchors most relevant to its cues and retrieve the
corresponding source evidence.

Specifically, we propose \textsc{EdgeMem}, a structured agent-memory method
with LLM-free construction and retrieval for multi-session interaction
histories. \textsc{EdgeMem}
uses local NLP tools and transcript metadata to annotate each turn with entity
and lexical keys, time, episode membership, and within-episode order. These
annotations form a multi-anchor hypergraph with three components: the Time
Sub-Hypergraph organizes turns by calendar time, the Co-occurrence
Sub-Hypergraph connects turns through shared entity and lexical keys, and the
Episode Sub-Hypergraph preserves session membership and local context. For a
new query, locally extracted entity, lexical, and date cues activate matching
hyperedges. Episode retrieval supplies the contextual core, while
co-occurrence and time retrieval add complementary evidence under a fixed
budget. The resulting evidence pack contains the original source turns and is
passed to a single reader LLM for answer generation. Because anchor construction
and query routing use local tools and deterministic rules, only this final
reader invokes an LLM.

To evaluate the effectiveness of \textsc{EdgeMem}, we compare it on LoCoMo with
six reproducible end-to-end memory baselines: A-Mem, CAM, CompassMem, MemoryOS,
MemGAS, and SeCom. On LongMemEval-S, we compare with the benchmark's published
full-history, dense-retrieval, conversational-memory, and structured-memory
baselines. Under the shared readout protocol on LoCoMo, \textsc{EdgeMem}
achieves the highest reported judge-based and lexical scores, including 61.01
under the strict judge versus 58.70 for the strongest baseline. It also
correctly rejects 92.57\% of unsupported-premise questions, compared with
84.68\% for the strongest baseline under the same refusal instruction. On
LongMemEval-S, the same configuration obtains the best session-level retrieval
scores, with 81.49 Recall@3 and 90.49 NDCG@3, and reaches an answer score of
60.00 versus the best result of 60.20 while using only 36\% as many tokens per
question. These results demonstrate strong memory retrieval and
memory-grounded question answering at substantially lower generative-LLM cost.

In summary, this work makes three core contributions:
\begin{itemize}
    \item We introduce \textsc{EdgeMem}, a source-preserving agent-memory
    framework that organizes original interaction turns in a multi-anchor
    hypergraph, enabling memory construction and retrieval without
    generative-LLM calls.

    \item We design a deterministic, budget-aware retrieval policy that uses
    episode context as its evidence core and additively incorporates
    cross-session co-occurrence and time-scoped evidence from the corresponding
    Sub-Hypergraphs.

    \item Extensive experiments on LoCoMo and LongMemEval-S show that
    \textsc{EdgeMem} achieves the strongest answer quality under the shared
    readout on LoCoMo and the best session-level retrieval at $k=3$ on
    LongMemEval-S, while using substantially fewer generative-LLM tokens than
    competing systems.
\end{itemize}

\begin{figure*}[t]
\centering
\includegraphics[width=0.99\textwidth]{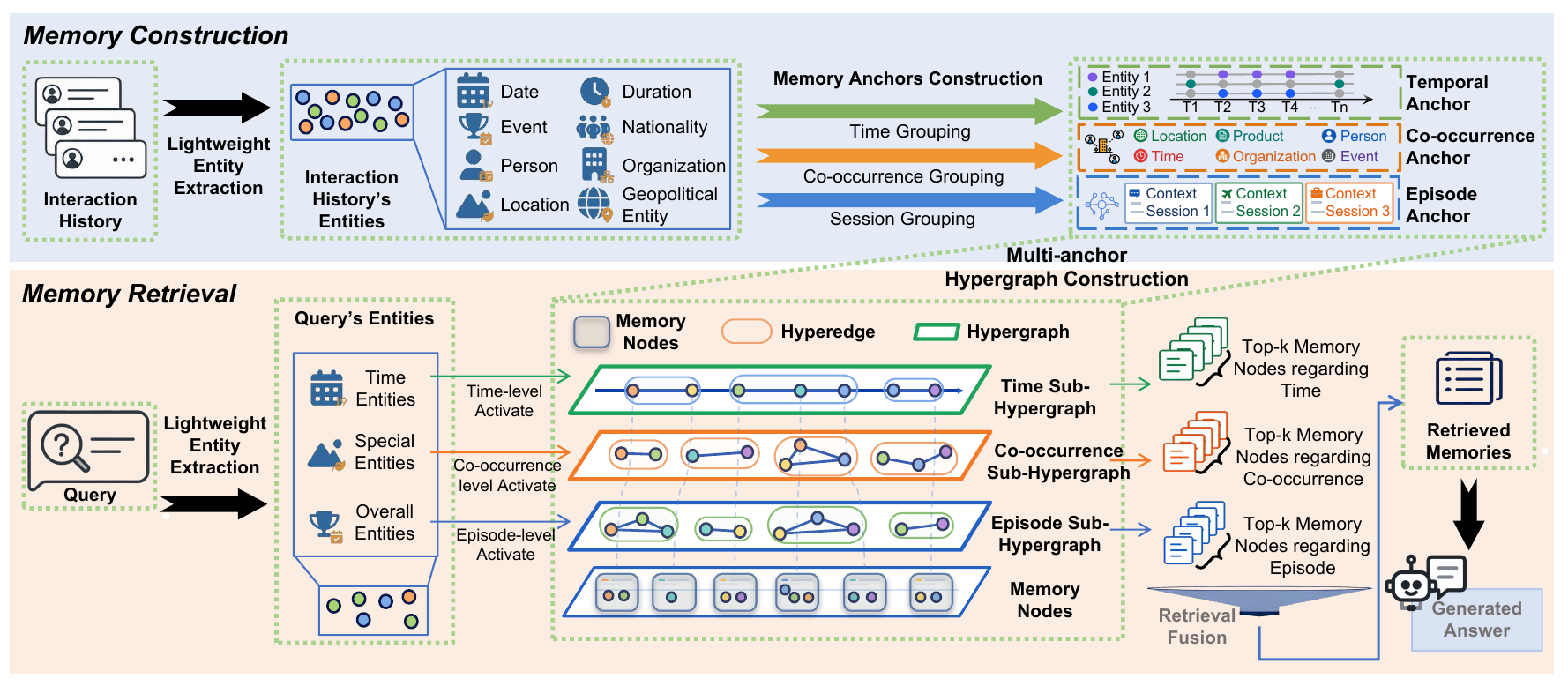}
\caption{\textsc{EdgeMem} overview. \emph{Top} (construction): the interaction
history is retained as source turns and processed by lightweight local
annotation; time, co-occurrence, and session grouping derive the time,
co-occurrence, and episode anchor keys of one multi-anchor hypergraph.
\emph{Bottom} (retrieval): query cues drive deterministic evidence channels
over the Time, Co-occurrence, and Episode Sub-Hypergraphs. Their selected
source turns are fused under a fixed budget and passed to a single reader,
which returns an answer or an explicit refusal. The reader is the pipeline's
single LLM stage.}
\Description{A two-stage diagram. During memory construction, source turns are
retained unchanged and locally annotated with entities, lexical content,
episode membership, and dates. Time, shared-key, and session grouping induce
the Time, Co-occurrence, and Episode Sub-Hypergraphs within one multi-anchor
hypergraph. During retrieval, date, entity, and lexical query cues drive the
three deterministic evidence channels. Their source turns are fused into one
evidence pack, after which a single language-model reader produces an answer
or refusal.}
\label{fig:framework}
\end{figure*}

\section{Related Work}
\label{sec:related}

Prior work relevant to agent memory spans three lines:
LLM-written conversational memory, graph-organized memory and retrieval, and
source-level retrieval informed by episodic models of contextual recall.

\paragraph{LLM-written conversational memory.}
A dominant line uses a language model at write time to summarize turns into
linked notes \citep{xu2025amem}, distill observations into hierarchical stores
\citep{kang2025memoryos}, consolidate facts through extract-and-merge policies
\citep{chhikara2025mem0,zhong2024memorybank}, recursively summarizing the
history \citep{wang2025recursum}, or compress topically segmented units
\citep{pan2025secom}. More recently, HyperMem organizes LLM-derived topics,
episodes, and facts in a hierarchical hypergraph and performs coarse-to-fine
retrieval \citep{yue2026hypermem}; although its episode nodes retain raw
dialogue, episode segmentation, topic aggregation, fact extraction, and
importance weighting still rely on LLM inference at write time. Such
representations require model-based construction as histories are processed;
details omitted from them may be unavailable to later questions.

\paragraph{Graph-organized memory and retrieval.}
Graph-based methods organize stored knowledge for traversal or diffusion. They
represent memories as knowledge-graph triples with personalized-PageRank
retrieval \citep{gutierrez2025hipporag2}, hierarchical summary graphs navigated
by an LLM \citep{li2025cam}, event graphs searched agentically
\citep{hu2026compassmem}, or multi-granularity graphs with PPR fusion
\citep{xu2026memgas}. LinearRAG instead constructs a graph for document
retrieval without LLM relation extraction \citep{zhuang2025linearrag}, showing
that graph organization need not depend on generative construction. In many
agent-memory systems, however, graph construction or query-time navigation
still introduces LLM calls.

\paragraph{Source-level retrieval and episodic organization.}
Source-level retrieval offers a complementary foundation. Retrieval-augmented
generation grounds answers in selected source passages \citep{lewis2020rag},
and BM25 ranks stored text without generation \citep{robertson2009bm25}.
Applied directly to dialogue, flat relevance by itself treats turns
independently and leaves relations such as shared events, episode membership,
and time implicit. Cognitive accounts of episodic memory instead characterize
an occurrence through both its content and spatiotemporal context
\citep{tulving1972episodic,clayton1998episodic}, motivating memory structures
that preserve source records while exposing multiple contextual access paths.

Overall, prior work offers complementary strengths in generated abstraction,
graph association, and direct source retrieval. Yet generation-based memory
introduces LLM cost and makes evidence fidelity depend on intermediate
representations, while flat source retrieval does not by itself encode temporal
and episodic structure. These properties are not typically combined in one
agent-memory design. Section~\ref{sec:method} introduces \textsc{EdgeMem}, which
preserves original turns and organizes LLM-free construction and retrieval
through locally derived time, co-occurrence, and episode anchors.

\section{\textsc{EdgeMem}: LLM-Free Structured Memory Management}
\label{sec:method}

We propose \textsc{EdgeMem}, a source-preserving agent-memory method for
LLM-free memory construction and retrieval. Figure~\ref{fig:framework} shows
the overall multi-anchor hypergraph pipeline.

\subsection{Problem Statement}
\label{sec:problem}

We study agent-memory answering over a multi-session interaction history,
represented as an ordered sequence of source turns
$\mathcal{T}=(t_1,\ldots,t_N)$. A turn
$t=(x_t,u_t,s(t),i(t),\tau_t)$ contains its original text $x_t$, transcript
speaker $u_t$, native session $s(t)$, within-session position $i(t)$, and
session date $\tau_t$. We call each native session an \emph{episode} and write
$T_s=(t_{s,1},\ldots,t_{s,|T_s|})$ for its ordered turns. Speaker identity
provides reader-facing provenance, while retrieval anchors are derived from record
content and structure.

A memory system is a triple $(\Phi,\Psi,F)$. Memory construction
$\Phi(\mathcal{T})$ produces a store $\mathcal{M}$; memory retrieval
$\Psi(q,\mathcal{M})$ maps a query $q$ to an evidence pack $P(q)$; and the
answer-stage language model $F$, termed the \emph{reader}, maps $(q,P(q))$ to a
response. The pack contains at most $\kappa$ source turns and $B$ rendered
tokens. The separation gives each stage a clear role: $\Phi$ controls
addressability, $\Psi$ fills the fixed evidence budget, and $F$ composes the
answer from that evidence.

\subsection{Local Annotation for Anchor Keys}
\label{sec:annotation}

To provide the anchor keys used in memory construction, we store every source
turn unchanged and derive the local annotations shown in
Figure~\ref{fig:framework} using local NLP tools. This gives \textsc{EdgeMem}
the design invariant
\begin{equation}
\mathrm{tok}_{\mathrm{LLM}}(\Phi)=
\mathrm{tok}_{\mathrm{LLM}}(\Psi)=0.
\label{eq:zerotoken}
\end{equation}
Entity and lexical keys expose what a turn discusses, while session and date
metadata preserve its coordinates in the interaction history. Local
construction and retrieval realize $\Phi$ and $\Psi$; the reader $F$ is reserved for
answer composition.\footnote{The reported implementation uses spaCy~3.8.11 with
\texttt{en\_core\_web\_sm}, \texttt{rank\_bm25}~0.2.2, and
\texttt{dateparser}~1.2.2.}

For each source turn, the named-entity set $E_t$ retains surface spans labeled
as person, organization, geopolitical entity or location, nationality or
group, named event, product, or date/time. The keyword set $K_t$ contains
lemmatized content tokens after function words, determiners, and possessives
are removed. These surface annotations keep anchor formation tied to observable
text and metadata. The session date $\tau_t$ comes directly from transcript
metadata. A date/time mention inside $x_t$ contributes a shared content key,
while $\tau_t$ supplies the record's calendar position.

The stored record retains $(x_t,u_t,s(t),i(t),\tau_t)$ together with $E_t$ and
$K_t$, with $x_t$ preserved verbatim. Speaker metadata is serialized for the
reader, and anchor membership is derived from $E_t$, $K_t$, $\tau_t$, and
$s(t)$. Together, the fields make each source turn addressable by content,
session, and time.

\subsection{Multi-anchor Hypergraph Construction}
\label{sec:hypergraph}

To make each source turn accessible by time, shared content, and session
context, we connect the annotated turns to three anchor-key families in one
multi-anchor hypergraph. Formally, $H=(V,\mathcal{E})$ has source turns as
vertices, $V=\{t_1,\ldots,t_N\}$, and typed hyperedges
$\mathcal{E}=\mathcal{E}_{\mathrm{time}}\sqcup
\mathcal{E}_{\mathrm{co}}\sqcup\mathcal{E}_{\mathrm{epi}}$:
\begin{equation}
e^{\mathrm{time}}_m=\{t\mid\tau_t\in m\},\qquad
e^{\mathrm{co}}_w=\{t\mid w\in E_t\cup K_t\},\qquad
e^{\mathrm{epi}}_s=T_s.
\label{eq:hubs}
\end{equation}
Hyperedges fit this many-to-many structure: turns expose several cues, and each
cue can connect evidence across episodes.
An anchor key therefore induces one hyperedge, or \emph{hub}, containing all
source turns that share that key. A source-turn record is stored once even
when it belongs to several hubs.

The three grouping operations in Figure~\ref{fig:framework} instantiate these
hub families. \emph{Time Grouping} maps the transcript date to a calendar-month
key $m$. \emph{Co-occurrence Grouping} creates a hub for each entity or content
key $w$; its document frequency $\mathrm{df}(w)$ counts incident source turns.
\emph{Session Grouping} treats each native session as one episode and retains
the order of its turns. The resulting Time, Co-occurrence, and Episode
Sub-Hypergraphs support date-span access, shared-key access, and episode-level
context, respectively.

Construction is append-only: each incoming turn contributes one stable source
record and links it to its time, co-occurrence, and episode hubs, allowing the
three access structures to grow together.

\subsection{Memory Retrieval}
\label{sec:retrieval}

To retrieve complementary evidence for a query, we extract its \emph{Date
Span}, \emph{Entity Cues}, and \emph{Lexical Cues} and route them through the
corresponding Time, Co-occurrence, and Episode Sub-Hypergraphs in
Figure~\ref{fig:framework}. Local parsers and lexical scorers execute the three
channels deterministically. Episode supplies the contextual core,
co-occurrence links recurring content across sessions, and time focuses an
explicit interval.

\paragraph{Query Cue Extraction.}
The local annotation pipeline produces $E_q$ and $K_q$ for the query. A fixed
normalization removes leading determiners and possessives from noun chunks;
$A_q=E_q\cup K_q$ forms the \emph{Entity Cues} used for shared-key matching.
The tokenized query forms the \emph{Lexical Cues}. Two BM25 indices
\citep{robertson2009bm25} score the original turn text $x_t$ and the
concatenated episode text $X_s=\mathop{\Vert}_{t\in T_s}x_t$, yielding
$r_{\mathrm{turn}}(q,x_t)$ and $r_{\mathrm{epi}}(q,X_s)$. A guarded parser
returns a \emph{Date Span} $D_q$ or $\bot$.

\paragraph{Episode Channel.}
The Episode channel builds the main evidence core by combining turn-level and
episode-level lexical relevance:
\begin{equation}
\rho(t)=\lambda\,\widehat r_{\mathrm{turn}}(q,x_t)
+(1-\lambda)\,\widehat r_{\mathrm{epi}}(q,X_{s(t)}).
\label{eq:score}
\end{equation}
Each score family is independently max-normalized and set to zero when all of
its scores are zero. The channel ranks turns by $\rho$, mixes the global
ranking with representatives from distinct episodes, and uses a fixed lexical
trigger to increase episode coverage for enumerative queries. Each selected
seed is read out with its immediate predecessor and successor when available.
Stable deduplication preserves the first occurrence of each turn, and the
channel stops after $k$ turns.

\paragraph{Co-occurrence Channel.}
The Co-occurrence channel applies the figure's \emph{Shared-key Match}. It
activates query keys whose document frequency lies in the rare-key band and
selects previously unseen source turns incident to those hubs:
\begin{align}
R_{\mathrm{co}}(q)
  &=\{w\in A_q\mid
    \delta_{\min}\leq\mathrm{df}(w)\leq\delta_{\max}\},\nonumber\\
C_{\mathrm{co}}(q)
  &=\operatorname*{top}_{\rho(t)}^{\,b_{\mathrm{co}}}
    \{t\notin C_{\mathrm{epi}}(q)\mid
      (E_t\cup K_t)\cap R_{\mathrm{co}}(q)\neq\emptyset\}.
\label{eq:cochannel}
\end{align}
The rare-key band favors anchors that recur enough to connect evidence while
remaining specific to a person, place, event, or topic. Because a hub may span
sessions, its highest-ranked turns can contribute evidence from another
episode. These turns fill the reserved Co-occurrence budget after the Episode
core.

\paragraph{Time Channel.}
The Time channel applies the figure's \emph{Date-span Gate}. It opens only when
$D_q$ contains an explicit month and four-digit year, retaining day precision
when present. Month hubs identify episodes inside the span; the fixed widening
rule also includes the first episode after the span when available. Candidate
turns are the episode openers and the highest-$\rho$ turns in this pool.
Stable deduplication admits at most $b_{\mathrm{time}}$ turns. When the gate is
closed, the channel is empty.

\paragraph{Retrieval Fusion and Evidence Pack.}
The Episode core is packed first. Co-occurrence evidence and, when activated,
Time evidence fill the remaining positions:
\begin{align}
b_{\mathrm{time}}&=\mathbf{1}[D_q\neq\bot]n_{\mathrm{time}},\nonumber\\
b_{\mathrm{co}}&=\min\!\left(n_{\mathrm{co}},
\max(0,\kappa-k-b_{\mathrm{time}})\right),\nonumber\\
P(q)&=\operatorname{dedup}\!\left(
C_{\mathrm{epi}}^{k}(q)\Vert
C_{\mathrm{co}}^{b_{\mathrm{co}}}(q)\Vert
C_{\mathrm{time}}^{b_{\mathrm{time}}}(q)\right).
\label{eq:channels}
\end{align}
Under the reported $k=14$, $n_{\mathrm{co}}=4$,
$n_{\mathrm{time}}=2$, and $\kappa=18$, the final four positions contain four
Co-occurrence turns when the Date-span Gate is closed, or two Co-occurrence
and two Time turns when it is open. Stable deduplication and the limits
$|P(q)|\leq\kappa$ and $\mathrm{tokens}(P(q))\leq B$ produce the
\emph{Evidence Pack} in Figure~\ref{fig:framework}.

\begin{table*}[t]
\centering
\caption{Comparison with released memory systems on LoCoMo, all numbers
from our metered runs: \textbf{(a)}~memory lifecycle cost,
\textbf{(b)}~answer quality under the shared answer prompt
(Sections~\ref{sec:setup} and~\ref{sec:positioning}). In this and the
following tables, the best value in each column is bold, the second best
underlined, arrows mark the preferred direction, and n/a marks quantities the
design does not define.}
\label{tab:grand}
\small
\par\centering \emph{\textbf{(a) Memory lifecycle cost, as released} --- LoCoMo without adversarial questions}\par\vspace{2pt}
\begin{tabular*}{0.96\textwidth}{@{\extracolsep{\fill}} l rrrrrr}
\toprule
System & \makecell[r]{Build\\tokens $\downarrow$} & \makecell[r]{Build\\calls $\downarrow$} & \makecell[r]{Build\\time $\downarrow$} & \makecell[r]{Tokens per\\question $\downarrow$} & \makecell[r]{Calls per\\question $\downarrow$} & \makecell[r]{Time per\\question (s) $\downarrow$} \\
\midrule
\textbf{EdgeMem} & \textbf{0} & \textbf{0} & \textbf{1.8\,min} & \textbf{1{,}000} & \textbf{1} & \textbf{1.3} \\
CAM & 1.27M & 1{,}516 & 75\,min & 19{,}745 & 4.9 & 11.8 \\
CompassMem & 1.76M & 1{,}500 & 43\,min & 41{,}348 & 21.7 & 45.8 \\
MemGAS & \underline{0.46M} & 544 & \underline{17\,min} & 3{,}146 & \underline{2} & 3.7 \\
MemoryOS & 4.49M & 12{,}730 & $\ge$1.6\,h & 5{,}116 & \underline{2} & 4.7 \\
A-Mem & 11.77M & 17{,}768 & 10.1\,h & 5{,}857 & \underline{2} & 5.2 \\
SeCom & 0.57M & \underline{310} & $\approx$1.9\,h$^\dagger$ & \underline{1{,}061} & \underline{2} & \underline{2.1} \\
\bottomrule
\end{tabular*}

\medskip
\par\centering \emph{\textbf{(b) Shared answer prompt} --- LoCoMo without adversarial questions}\par\vspace{2pt}
\begin{tabular*}{0.96\textwidth}{@{\extracolsep{\fill}} l rrrrrr}
\toprule
System & SJ $\uparrow$ & LJ $\uparrow$ & F1 $\uparrow$ & BLEU-4 $\uparrow$ & ROUGE-L $\uparrow$ & BERTScore $\uparrow$ \\
\midrule
\textbf{EdgeMem} & \textbf{61.01} & \textbf{77.86} & \textbf{22.44} & \textbf{5.03} & \textbf{21.82} & \textbf{85.90} \\
CompassMem & \underline{58.70} & \underline{77.01} & \underline{20.56} & 4.41 & \underline{19.81} & \underline{85.57} \\
CAM & 54.94 & 75.26 & 18.51 & 3.95 & 18.24 & 85.17 \\
MemoryOS & 47.20 & 67.73 & 16.23 & 3.11 & 15.81 & 84.94 \\
MemGAS & 44.05 & 61.69 & 19.50 & \underline{4.57} & 19.44 & 85.44 \\
A-Mem & 39.74 & 59.35 & 15.33 & 2.94 & 15.01 & 84.62 \\
SeCom & 38.12 & 52.99 & 14.70 & 3.10 & 14.54 & 84.49 \\
\bottomrule
\end{tabular*}

\end{table*}

\subsection{Generated Answer}
\label{sec:answering}

To produce a grounded response, we serialize the Evidence Pack and pass it
with the query to a single reader call $F$. Each evidence item retains the
transcript speaker, session date, and unchanged source text, supporting
provenance and temporal interpretation. The reader is the pipeline's single
generative stage and runs at temperature zero with a maximum of 128 output
tokens.

The evidence-only prompt in Appendix~\ref{sec:appx-reader} requests a short
answer, requests all items for a list question, and interprets relative dates
against the rendered session date. If the Evidence Pack is insufficient, the
reader returns the fixed response \texttt{Not mentioned in the conversation.}
This explicit outcome supports direct measurement of abstention behavior
against the retrieved pack~$P(q)$.

\section{Experiments}
\label{sec:experiments}

We design the evaluation to answer five questions. \textbf{Q1
(Question-Answering Accuracy)}: does removing generative inference from memory
construction and retrieval compromise answer quality? \textbf{Q2 (Memory
Management Efficiency)}: how efficiently does each system construct and use
its memory, in terms of both token usage and time? \textbf{Q3
(Original-Evidence Preservation)}: when the history does not support an answer,
how reliably does a source-preserving memory abstain? \textbf{Q4
(Anchor Ablation Study)}: what does each anchor contribute, alone and in
combination? \textbf{Q5 (Cross-Benchmark Generalization)}: does the same chain
transfer unchanged to a second benchmark? An audited case study closes the
section by connecting the aggregate gaps to their mechanism.

\subsection{Experimental Setup}
\label{sec:setup}

\paragraph{Datasets.}
LoCoMo \citep{maharana2024locomo} contains ten synthetic multi-session dialogues
(averaging 588.2 turns over up to 32 sessions) with 1{,}986 questions in five
categories, of which 446 are adversarial (category 5): the conversation does not
support the premised answer. LongMemEval-S \citep{wu2025longmemeval} contains
500 questions, each over its own haystack of roughly 50 sessions
($\sim$115k tokens); 30 questions are unanswerable by construction. We use the
2025-09 cleaned LongMemEval release (identical questions, lightly edited
haystacks).

\paragraph{Protocol.}
All components are frozen before evaluation: the reader is
\texttt{gpt-4o-mini-2024-07-18} at temperature 0, with the reader prompt
reproduced in Appendix~\ref{sec:appx-reader}; packed evidence is capped at $\kappa=18$ items and
$B=3{,}072$ tokens on both benchmarks. The remaining retrieval constants of
Section~\ref{sec:method} are likewise fixed at design time and never tuned:
Episode Sub-Hypergraph output budget $k=14$; mixing coefficient $\lambda=0.5$; diversity
floor $K=3$, raised to $5$ for enumerative questions; Co-occurrence
Sub-Hypergraph capacity $n_{\mathrm{co}}=4$ over the rare-hub band
$[\delta_{\min},\delta_{\max}]=[2,10]$; Time Sub-Hypergraph capacity
$n_{\mathrm{time}}=2$. Retrieval executes at \PNinetyFiveLatency{}\,ms P95 latency on a
desktop CPU. Every LLM call in every run --
ours and every reproduced baseline -- is metered per call and locked to the
same snapshot; a serving-side substitution is detected and retried, and runs
report zero unrepaired violations.

\paragraph{Metrics.}
Answer accuracy is graded by two LLM judges: a \emph{strict} judge (SJ),
which credits a reply only when it contains the reference answer, and a
\emph{lenient} judge (LJ), which also credits replies on the same topic or in
the same time period; both prompts follow prior evaluation practice and are
reproduced from released artifacts for cross-paper comparability
(both are reproduced in Appendices~\ref{sec:appx-4oj} and~\ref{sec:appx-lmj}). Alongside the judges we report F1, BLEU-4 \citep{papineni2002bleu},
ROUGE-L \citep{lin2004rouge} and BERTScore \citep{zhang2020bertscore}, and measured prompt$+$completion tokens per question. Lexical metrics are
comparable only within a generation style (Section~\ref{sec:positioning});
judge and BERTScore columns carry cross-style comparisons. For adversarial
questions we report three calibers: an exact-refusal rule (a reply is credited only
when it is the designated refusal sentence, reproduced exactly; 444 items) plus the judge on the 2 answerable
items, a lenient substring detector that coincides with LoCoMo's own
released category-5 rule, and a matched-instruction arm that transplants our
refusal sentence into each baseline's own prompt with nothing else changed. On LoCoMo, compared systems store heterogeneous memory
units---generated notes, compressed segments, raw turns---so retrieval has no
common unit, and memory quality is measured end to end through answer quality
under the shared budget. LongMemEval-S labels supporting evidence at the
session level and its suite reports session-level retrieval, giving a
benchmark-defined unit every system shares; we therefore additionally report
Recall and NDCG there, a measurement of the memory alone, upstream of any
reader or answer style. Statistical claims use paired conversation-cluster
bootstrap (10 clusters, 10{,}000 draws); bracketed ranges throughout the paper
are 95\% bootstrap confidence intervals.

\paragraph{Baselines.}
We compare against systems that satisfy one requirement: a released pipeline
that runs end to end under locked model snapshots, so that every number in
Table~\ref{tab:grand} is measured rather than quoted. Six systems meet it and
are reproduced from their released code, pinned and audited -- A-Mem
\citep{xu2025amem}, CAM \citep{li2025cam}, CompassMem \citep{hu2026compassmem},
MemoryOS \citep{kang2025memoryos}, MemGAS \citep{xu2026memgas} and SeCom
\citep{pan2025secom} -- together spanning the design space of
Section~\ref{sec:related}: generative note-writing, hierarchical stores,
multi-granularity graphs, compression-based construction, and query-time LLM
retrieval control. Systems that cannot be rerun this way enter comparison only
as transcribed rows marked $^\dagger$. On LongMemEval-S the comparison
follows the benchmark's published evaluation suite: full-history prompting, the
dense retrievers MPNet \citep{song2020mpnet} and Contriever
\citep{izacard2022contriever}, the conversational methods MPC
\citep{lee2023mpc}, RecurSum \citep{wang2025recursum} and SeCom, and the
memory systems A-Mem, HippoRAG~2 and MemGAS itself, with our rows evaluated
under the same protocol. The two
comparison sets thus differ by construction rather than by selection: LoCoMo
compares end-to-end memory systems, where cost and answer behavior are the
measurable quantities, while the LongMemEval-S suite centers on methods whose
retrieval is measurable on the shared session unit. Each
reproduced system appears both as released and under a fixed \emph{shared answer
prompt} (reproduced in Appendix~\ref{sec:appx-sharedprompt}), separating memory quality
from answer style. Reproduction
deviations, audits and per-run cost ledgers are released with the code.

\subsection{Question-Answering Accuracy (Q1)}
\label{sec:positioning}

Removing generative inference from memory construction and retrieval costs
no measurable answer quality; once answer style is controlled, the anchored
memory is the strongest in the comparison.

As released---each system under its own answer prompt and decoding, one judge
snapshot, one denominator---the three leading systems are statistically
indistinguishable under the strict judge: the
paired difference is $-0.65$ [$-2.88$, $+1.50$] against CAM and $+0.45$
[$-2.61$, $+3.30$] against CompassMem.

Holding answer style fixed sharpens the comparison. Panel (b) reads each
system's own retrieval evidence out through the shared answer prompt
(Section~\ref{sec:setup}), so differences between rows are attributable to
the memory rather than to phrasing conventions, and the lexical
metrics---F1, BLEU-4, ROUGE-L, BERTScore---become comparable across rows.
Under this control \textsc{EdgeMem} attains the highest value in every
column, and the ordering of the seven rows is identical under both judges
(agreement statistics in Appendix~\ref{sec:appx-lmj}), so the conclusion does
not depend on the evaluator. Why the anchored memory wins is the subject of
the ablation (Q4), which attributes the gap to anchor structure rather than
to the reader: accuracy is not the price of removing generative
inference---the anchored memory is the stronger one.

\subsection{Memory Management Efficiency (Q2)}
\label{sec:efficiency}

The accuracy of the preceding section is reached at the lowest token and
time cost in the comparison: construction is token-free, and the two systems
closest in accuracy spend one to two orders of magnitude more tokens per
question. Across the panel, quality does
not track spend: the most expensive system is not the strongest, and the
strongest memory is also the cheapest (Figure~\ref{fig:respanels}a).

In tokens, \textsc{EdgeMem} builds its store without any model calls and
answers within roughly one thousand tokens per question, where the two
systems it ties with spend 19{,}745 and 41{,}348 (Table~\ref{tab:grand}a).

In time, the same separation holds. Because the reproductions ran at
different worker counts, both figures are concurrency-independent: summed
per-call latency for builds, and per-question sums of the calls each
navigating system issues sequentially by design. Building
\textsc{EdgeMem}'s store is a 1.8-minute single-thread annotation pass on a
desktop CPU, against 17 minutes to 10.1 hours of API time for the metered
systems; a question is answered in about 1.3\,s end to end (10.73\,ms P95
retrieval), against 3.7--45.8\,s for systems that consult a model on the
query path.

Local NLP and embedding computation runs on CPU and is not billed. Timing
provenance per system: A-Mem's build is inherently sequential (its evolution
rewrites earlier notes; 17{,}768 calls at 2.0\,s mean, zero failures);
CompassMem's per-question time comes from its agentic run ($n{=}584$),
MemoryOS's from recorded per-question elapsed time, and SeCom's from a
thirty-question replay under the locked snapshot, with its build reported as
wall-clock ($^\dagger$). Zero-token organization thus turns the memory bill
from a recurring cost into a fixed, local one.

\subsection{Original-Evidence Preservation (Q3)}
\label{sec:adversarial}

\begin{table}[t]
\centering
\caption{Behavior on the 446 LoCoMo adversarial questions under the three
scoring arms of Section~\ref{sec:setup}.}
\label{tab:adv}
\footnotesize
\setlength{\tabcolsep}{4pt}
\begin{tabular}{l rrrr}
\toprule
System & \makecell[r]{Judge\\(all questions) $\uparrow$} & \makecell[r]{Exact\\refusal $\uparrow$} & \makecell[r]{Lenient\\refusal $\uparrow$} & \makecell[r]{Matched\\refusal $\uparrow$} \\
\midrule
\textbf{EdgeMem} & \textbf{61.76} & \textbf{92.57} & \textbf{92.57} & \textbf{92.57} \\
CAM & \underline{41.64} & n/a & 1.58 & 38.96 \\
CompassMem & 40.81 & n/a & 18.02 & 79.28 \\
MemGAS & 34.26 & n/a & \underline{23.65} & 74.77 \\
MemoryOS & 31.37 & n/a & 12.61 & 75.00 \\
A-Mem & 25.19 & n/a & 2.93 & \underline{84.68} \\
SeCom & 31.37 & n/a & 12.39 & 75.00 \\
\bottomrule
\end{tabular}
\end{table}

When the conversation does not support the premised answer,
\textsc{EdgeMem} abstains far more reliably than any baseline given the same
instruction; the price it pays on the answerable split is measured and
reported.

Table~\ref{tab:adv} scores the 446 adversarial questions under three
arms declared in advance: \emph{exact} credits only the designated refusal
sentence, reproduced exactly (n/a for systems whose released prompts carry no refusal
instruction; measured at zero for every such row); \emph{lenient} coincides with the detector in LoCoMo's own
released evaluation and credits any phrasing that declares the information
absent; \emph{matched} transplants our refusal sentence into each baseline's
own prompt with nothing else changed, and is therefore the arm that compares
ability rather than the presence of the instruction.

Under the matched arm every
baseline improves substantially, yet none reaches \textsc{EdgeMem}'s native
92.57: five of the six cluster between 74.8 and 84.7, and CAM remains at
38.96---its reader, supplied with roughly twenty thousand tokens of summarized
context, continues to find apparent support for the premised answer. The
instruction alone does not account for our margin: appending the same sentence
to the shared answer prompt of panel (b) yields 76.35, so part of the remaining gap
is carried by the short-answer output format.

The evidence-only rule trades answerable-split recall for
reliability on unsupported premises. The same conservatism carries a
measurable cost on the answerable split: the reader abstains on 28.0\%
of the 1{,}540 answerable questions, and the recoverable share is about eight
judge points, since the unconstrained shared answer prompt of panel (b) scores 61.01 on
identical evidence. The all-set column reports the net effect of this
trade-off. Reliable abstention, then, is a property of evidence-only
answering over preserved sources, not of the instruction alone.

\subsection{Anchor Ablation Study (Q4)}
\label{sec:ablation}

\begin{figure}[t]
\centering
\begin{minipage}[b]{0.49\columnwidth}\centering
\includegraphics[width=\linewidth]{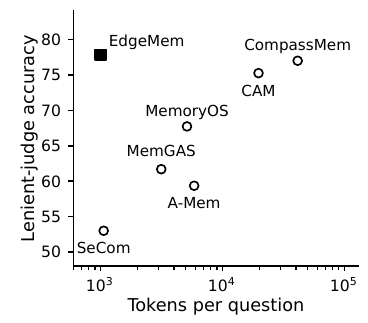}
{\small (a)}
\end{minipage}\hfill
\begin{minipage}[b]{0.49\columnwidth}\centering
\includegraphics[width=\linewidth]{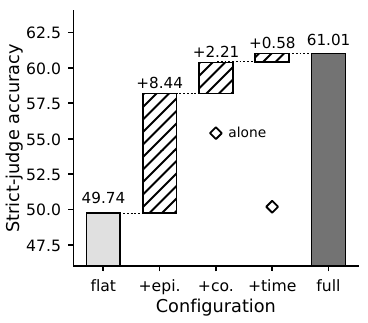}
{\small (b)}
\end{minipage}
\caption{(a) Memory quality (lenient judge, Table~\ref{tab:grand}b) against
each released pipeline's per-question token cost (Table~\ref{tab:grand}a);
upper left is better. (b) The anchor ladder: the accuracy each anchor adds in
deployment order (Table~\ref{tab:anchorladder}); diamonds show co-occurrence and
time alone.}
\Description{Two small panels. Left: scatter of accuracy versus token cost on
a logarithmic axis; EdgeMem sits in the upper left. Right: waterfall from a
flat baseline through episode, co-occurrence, and time increments to the full chain,
with diamonds for co-occurrence-only and time-only variants.}
\label{fig:respanels}
\end{figure}

\begin{table}[t]
\centering
\caption{Anchor ladder on LoCoMo under the shared answer prompt: the
deployed path adds one Sub-Hypergraph per rung (upper block); the lower block
lists non-deployed single- and pair-anchor variants.}
\label{tab:anchorladder}
\footnotesize
\setlength{\tabcolsep}{4.5pt}
\begin{tabular}{l ccccc}
\toprule
Configuration & SJ $\uparrow$ & F1 $\uparrow$ & BLEU-4 $\uparrow$ & ROUGE-L $\uparrow$ & BERTScore $\uparrow$ \\
\midrule
flat turn BM25 & 49.74 & 19.40 & 4.14 & 19.00 & 85.38 \\
\makecell[l]{$+$ episode\\Sub-Hypergraph} & 58.18 & 21.94 & 4.88 & 21.41 & 85.80 \\
\makecell[l]{$+$ co-occurrence\\Sub-Hypergraph} & \underline{60.43} & 22.12 & 4.92 & 21.53 & 85.83 \\
\makecell[l]{$+$ time\\Sub-Hypergraph} & \textbf{61.01} & \textbf{22.44} & \textbf{5.03} & \textbf{21.82} & \textbf{85.90} \\
\midrule
co-occurrence only & 55.39 & 20.66 & 4.52 & 20.07 & 85.64 \\
time only & 50.19 & 19.54 & 4.19 & 19.16 & 85.40 \\
episode $+$ time & 58.96 & \underline{22.23} & \underline{4.97} & \underline{21.69} & \underline{85.87} \\
\bottomrule
\end{tabular}
\end{table}

\begin{table}[t]
\centering
\caption{LongMemEval-S: session-level Recall (R) and NDCG (N) at
$k\in\{3,10\}$ ($n{=}470$), strict-judge accuracy (SJ), and reader input
tokens per question.}
\label{tab:lme}
\small
\setlength{\tabcolsep}{3pt}
\begin{tabular}{l cc cc c r}
\toprule
& \multicolumn{2}{c}{@3} & \multicolumn{2}{c}{@10} & & \\
\cmidrule(lr){2-3}\cmidrule(lr){4-5}
System & R $\uparrow$ & N $\uparrow$ & R $\uparrow$ & N $\uparrow$ & SJ $\uparrow$ & \makecell[r]{Tokens per\\question $\downarrow$} \\
\midrule
Full History$^\dagger$ & n/a & n/a & n/a & n/a & 50.60 & 103{,}137 \\
MPNet$^\dagger$ & 66.17 & 75.47 & 85.11 & 80.63 & 53.20 & 8{,}173 \\
Contriever$^\dagger$ & 71.06 & 79.72 & 90.00 & 84.29 & 55.40 & 8{,}286 \\
MPC$^\dagger$ & 60.00 & 70.90 & 80.00 & 76.59 & 53.80 & 8{,}457 \\
RecurSum$^\dagger$ & 67.23 & 78.33 & 87.66 & 83.28 & 35.40 & 8{,}853 \\
SeCom$^\dagger$ & 71.06 & 80.88 & 89.15 & 85.11 & 56.00 & \textbf{2{,}741} \\
HippoRAG 2$^\dagger$ & 75.53 & 85.44 & 91.28 & 88.73 & \underline{57.60} & 8{,}530 \\
A-Mem$^\dagger$ & n/a & n/a & n/a & n/a & 55.60 & 9{,}018 \\
MemGAS$^\dagger$ & \underline{78.51} & \underline{86.83} & \textbf{94.47} & \underline{89.96} & \textbf{60.20} & 8{,}829 \\
\midrule
\textbf{EdgeMem} & \textbf{81.49} & \textbf{90.49} & \underline{93.83} & \textbf{92.80} & 60.00 & \underline{3{,}202} \\

\bottomrule
\end{tabular}
\vspace{-8pt}
\end{table}

\begin{table*}[t]
\centering
\caption{Three audited failures, each traced from memory through retrieval to
the answer, against \textsc{EdgeMem}'s handling of the same questions: a
faithful memory retrieved incompletely (A-Mem), a key fact destroyed during
summarization (CAM), and a faithful store whose agentic navigation selects no
evidence (CompassMem). \cmk: correct or preserved; \xmk: missed or incorrect.}
\label{tab:casestudy}
\footnotesize
\setlength{\tabcolsep}{5pt}
\begin{tabular*}{\textwidth}{@{\extracolsep{\fill}} l p{4.8cm} p{4.8cm} p{4.4cm}}
\toprule
 & Memory / summary & Retrieved evidence & Final answer \\
\midrule
\multicolumn{4}{@{}l}{\emph{Q1: What gaming equipment did John buy or
refurbish?}\hfill gold: \emph{Sennheiser headphones, Logitech mouse, gaming desk}}\\[2pt]
A-Mem & \cmk\ notes faithfully retain headphones, \okhl{mouse}, and the
refurbished \okhl{desk} & \xmk\ partial retrieval: D23:10 omitted;
\badhl{mouse and desk} never reach the reader & \xmk\ ``new devices and
headphones''---misses \badhl{mouse and desk} \\
\textbf{EdgeMem} & \cmk\ original turns kept: ``gaming desk'',
``headphones'', ``Sennheiser + Logitech mouse'' & \cmk\ \okhl{D20:9, D23:8,
D23:10} packed (all gold) & \cmk\ ``headphones, \okhl{mouse}, and refurbished
\okhl{desk}'' \\
\midrule
\multicolumn{4}{@{}l}{\emph{Q2: What kind of beer does McGee's bar
serve?}\hfill gold: \emph{stout, lager}}\\[2pt]
CAM & \xmk\ key fact dropped: gist layer omits \badhl{``McGee's'' and
``lager''}; the exact leaf is pruned & \xmk\ cannot recover what construction
discarded & \xmk\ ``stout and light beers''---misses \badhl{lager} \\
\textbf{EdgeMem} & \cmk\ original turns kept: D21:15 (great stout),
D23:3 (``good \okhl{lager} at McGee's'') & \cmk\ \okhl{D21:15, D23:3} packed &
\cmk\ ``great stout and \okhl{lager} beer'' \\
\midrule
\multicolumn{4}{@{}l}{\emph{Q3: Who was the new addition to Nate's family in
May 2022?}\hfill gold: \emph{Max}}\\[2pt]
CompassMem & \cmk\ faithful event node: ``Nate introduces his new dog,
\okhl{Max}'' & \xmk\ retrieval miss: agentic navigation selects \badhl{zero
nodes}; N10 never enters the reader & \xmk\ \badhl{``not mentioned''} \\
\textbf{EdgeMem} & \cmk\ original turn kept: D12:3 (``new addition to
the family, this is \okhl{Max}!'') & \cmk\ gold evidence \okhl{D12:3} packed &
\cmk\ \okhl{``Max''} \\
\bottomrule
\end{tabular*}
\end{table*}

Every anchor earns its place: each contributes a significant increment in
deployment order, episode structure carries the largest step, and no
combination of fewer anchors matches the deployed chain.
Table~\ref{tab:anchorladder} adds the three anchors one at a time
(Figure~\ref{fig:respanels}b) with the
answer style held fixed, so successive rows differ only in the memory; its
final configuration is the \textsc{EdgeMem} row of Table~\ref{tab:grand}(b). Flat
BM25 over raw turns at a 14-item budget reaches 49.74 under the judge
(effective $n$ = 1{,}540/1{,}539; one question filtered in the upper
configurations). Reading each matched
turn together with its episode-local context at the same budget raises the
score to 58.18 ($+$8.44 [5.79, 10.91]): most of what the memory contributes is
the surrounding episode rather than the matched turn alone. Extending the pack
with up to four strictly additive co-occurrence items adds $+$2.21
[0.76, 3.50], drawing in cross-episode evidence about the same people and
places without displacing the episode core. The guarded Time Sub-Hypergraph output swaps
only the final two positions of the pack under a deterministic date parser
and contributes a
further $+$0.58 [0.07, 1.15], concentrated on date-conditioned questions. All three intervals
exclude zero, every lexical column moves in the same direction, and the
evidence budget grows only from 727 to 944 tokens across the full ladder, so
the accuracy gains are attributable to anchor structure rather than to a
larger evidence pack.

Module-alone and pairwise variants confirm the ordering and
expose the episode--co-occurrence overlap. The lower block of
Table~\ref{tab:anchorladder} completes the picture with single- and pair-module variants, constructed for
ablation only: co-occurrence items seeded from the flat ranking, and the Time Sub-Hypergraph
transplanted under the same gate. Alone, the anchors rank in the same order as
their marginal contributions---episode $+$8.44, co-occurrence $+$5.65, time $+$0.45
over flat BM25. The pairwise cells explain why the later ladder increments are
smaller: episode and co-occurrence overlap (58.18 and 55.39 alone, 60.43 together,
about 3.4 points short of additive), since cross-episode co-occurrence items partly
substitute for episode context, and the Time Sub-Hypergraph output is worth more on the
episode base ($+$0.78) than on flat ($+$0.45)---its replacement turns are
episode openers, which the episode read-out is built to exploit. Every variant
is dominated by the deployed chain. Read along the alternative path---co-occurrence
first, then episode---the grid gives $49.74 \to 55.39 \to 60.43$, increments
of $+$5.65 and $+$5.04: the attribution between the two anchors shifts with
insertion order, as their overlap predicts, but the destination and the time
Time Sub-Hypergraph's small final step do not. The ladder, the grid, and both insertion
orders point to the same conclusion: the anchors, not generation, carry the
memory.

\subsection{Cross-Benchmark Generalization (Q5)}
\label{sec:lme}

Table~\ref{tab:lme} reports a transfer check on LongMemEval-S: nothing in
the memory or retrieval is retuned, and the comparison set and metric
implementation follow the benchmark's standard suite ($^\dagger$ rows in
Table~\ref{tab:lme} are transcribed from \citet{xu2026memgas}), since each
question carries its own haystack of roughly 115k tokens and rerunning
generative-construction systems at that scale multiplies their build cost by
500 haystacks. The unchanged chain
attains the highest session-level Recall@3 (81.49) and NDCG@3 (90.49); because
the deployed chain reads only about three sessions per question, the rank-3
column doubles as a small-budget test of the same structure. Under the style
control, which re-answers \textsc{EdgeMem}'s byte-identical evidence with the
shared answer prompt (Section~\ref{sec:setup}) and is therefore excluded from
the table's ranking, its strict-judge score reaches 60.00, close to
60.20 for the best transcribed system, on roughly a third of the evidence
tokens. On the 30 unanswerable questions, the house refusal rule abstains at
90.00\%. The structure transfers without retuning.

\subsection{Case Study: How Generated Memories Lose Evidence}
\label{sec:audit}

To connect the aggregate gaps to mechanisms, we audited 27 failures of the
reproduced systems end to end, tracing each question from the original turns
through construction and retrieval to the answer. The recurring mechanisms
are retrieval misses, dates dropped or shifted during extraction, noise
admitted by link expansion, and facts or qualifiers lost in summarization; a
single failure often exhibits several, because a detail altered at one stage
is consumed as ground truth by the next, and one system ships with its chain
broken outright: MemGAS's released generation never consumes its retrieved
evidence. Table~\ref{tab:casestudy} traces
three of them, one per failure surface.

In the first, A-Mem's notes faithfully retain the queried facts, but
note-level retrieval returns only part of them, and the answer misses two of
the three items. In the second, CAM's hierarchical summarization drops the
key fact while building its gist layer and prunes the leaf that contained it;
no retrieval strategy can recover what construction has discarded, so the
failure is decided at write time. In the third, CompassMem stores a faithful
event node, yet its agentic navigation selects no nodes for the reader, and
the system answers ``not mentioned'' while the evidence sits in its store.
\textsc{EdgeMem} answers all three questions from the unaltered turns: the
anchors route directly to the gold evidence, and nothing upstream has
rewritten or pruned it.

The three cases span the pipeline: memory that is faithful but incompletely
retrieved, memory that destroys evidence before any question is asked, and
retrieval that fails over a faithful store. Only the write-time loss is
unrecoverable in principle; the other two are the recurring price of
searching derived structures with further inference. A source-preserving
store with deterministic anchored retrieval proves effective against all
three failure surfaces in these audits, which substantiates the risk stated in
Section~\ref{sec:intro}: what generative construction alters or discards is
rarely recoverable downstream, whereas a store that leaves the evidence
untouched leaves resolution to the one component that sees the question and
the original turns together.

\section{Conclusion}
\label{sec:conclusion}

We introduce \textsc{EdgeMem}, an LLM-free agent memory framework that uses
lightweight local annotation over multi-session interaction histories. It
stores original interaction turns in a multi-anchor hypergraph with Time,
Co-occurrence, and Episode Sub-Hypergraphs built from the transcript. Upon this
structure, retrieval activates the hyperedges a
query matches and fuses the three Sub-Hypergraph outputs
into a single evidence pack, so that every answer is composed by one reader
call and remains traceable to the turns behind it. On LoCoMo, \textsc{EdgeMem}
attains the strongest answer quality under a shared prompt and abstains more
reliably than the compared baselines when the conversation does not support an
answer. The same fixed configuration obtains the highest session-level
Recall@3 and NDCG@3 in the reported suite. Across both benchmarks, memory
construction and retrieval use zero generative-LLM tokens, yielding efficient,
source-grounded agent memory.

\section*{Ethical Considerations}
We use two publicly released research benchmarks, LoCoMo and LongMemEval-S,
containing synthetic or curated conversations, and conduct no new data
collection or human-subject study. All datasets and reproduced systems are
used under their released licenses. Because \textsc{EdgeMem} preserves source
turns verbatim, real-world deployments should enforce consent-aligned
retention, access-control, and deletion policies; these deployment-level
mechanisms are outside the scope of our evaluation.

\bibliographystyle{ACM-Reference-Format}
\bibliography{references}

\clearpage
\appendix
\appendix
\section{Prompts and Scoring Rules}
\label{sec:appendix-prompts}

All prompts below are reproduced verbatim from the released artifacts. Source
identifiers and available SHA-256 checksums are reported with each prompt.

\subsection{Reader Prompt}
\label{sec:appx-reader}

The prompt of \textsc{EdgeMem}'s own reader
(\path{gpt-4o-mini-2024-07-18}, temperature 0), used in every native
\textsc{EdgeMem} run. Its final sentence is the \emph{abstention
instruction}: the exact refusal string it mandates is what the adversarial
rule checks for, and the matched-instruction arm transplants this sentence,
unmodified, into each baseline's own prompt.

\begin{promptbox}{Reader Prompt}
You answer questions about a long conversation using ONLY the evidence provided.\\
Dates: when the question asks when something happened, resolve relative expressions
(yesterday, last week, next month) against the [session\_date] of the evidence turn
and answer with the resolved calendar date.\\
Answer with a short phrase; when the answer is a list, include every item.\\
If relevant evidence is present, answer from it; only if the evidence does not
contain the answer, reply exactly: Not mentioned in the conversation.
\end{promptbox}

\subsection{Shared Answer Prompt}
\label{sec:appx-sharedprompt}

\begingroup\sloppy
The shared answer prompt of Table~\ref{tab:grand}(b) and of the
style-control row of Table~\ref{tab:lme} is the answer-stage prompt of
\citet{xu2026memgas}, lifted verbatim from their release (file
\path{src/generation/generation_multigran.py}, constant
\texttt{PROMPT\_G}, sha256 \texttt{22797554\ldots}). Decoding follows the
release: \texttt{gpt-4o-mini-2024-07-18}, temperature 0, max tokens
4{,}000. It carries no refusal or length constraint (the sense in which the
text calls it \emph{unconstrained}); each system's retrieved evidence enters
\texttt{\{retrieved\_texts\}} with nothing else changed.\endgroup

\begin{promptbox}{Shared Answer Prompt}
You are an intelligent dialog bot. You will be shown History Dialogs. Please
read, memorize, and understand the given Dialogs, then generate one concise,
coherent and helpful response for the Question.\\
History Dialogs: \{retrieved\_texts\}\\
Question Date: \{question\_date\}\\
Question: \{question\}
\end{promptbox}

\subsection{Strict Judge (SJ)}
\label{sec:appx-4oj}

The strict-judge columns (SJ) use the judge prompt of \citet{xu2026memgas} (their Appendix~J,
Fig.~12), reproduced verbatim and served by the pinned snapshot
\texttt{gpt-4o-2024-08-06} at temperature 0:

\begin{promptbox}{Strict Judge Prompt (SJ)}
I will give you a question, a reference answer, and a response from a model. Please answer [[yes]] if the response contains the reference answer. Otherwise, answer [[no]]. If the response is equivalent to the correct answer or contains all the intermediate steps to get the reference answer, you should also answer [[yes]]. If the response only contains a subset of the information required by the answer, answer [[no]].\\
{}[User Question]\\
\{question\}\\
{}[The Start of Reference Answer]\\
\{answer\}\\
{}[The End of Reference Answer]\\
{}[The Start of Model's Response]\\
\{response\}\\
{}[The End of Model's Response]\\
Is the model response correct? Answer [[yes]] or [[no]] only.
\end{promptbox}

\subsection{Lenient Judge (LJ)}
\label{sec:appx-lmj}

\begingroup\sloppy
The lenient-judge columns (LJ) use the LoCoMo judge prompt of LightMem \citep{fang2026lightmem},
lifted verbatim at run time from the pinned release (repository
\path{zjunlp/LightMem}, commit \texttt{8fc9a91}, key
\path{locomo-judge}, sha256 \texttt{406f2215\ldots}) and served by
\path{gpt-4o-mini-2024-07-18} at temperature 0 with the release's JSON
label contract. It is the Mem0-lineage ``generous'' judge: same-topic answers
and same-period time references count as correct; following that release,
adversarial questions are excluded from its denominator ($n{=}1{,}540$). On
20{,}013 paired verdicts, LJ upgrades 17.7\% of SJ's negatives and
downgrades 0.1\%, and the row orderings of Table~\ref{tab:grand}(b) are
identical under the two judges.\endgroup

\begin{promptbox}{Lenient Judge Prompt (LJ)}
Your task is to label an answer to a question as 'CORRECT' or 'WRONG'. You will be given the following data: (1) a question (posed by one user to another user), (2) a 'gold' (ground truth) answer, (3) a generated answer which you will score as CORRECT/WRONG.\\{}
\par
The point of the question is to ask about something one user should know about the other user based on their prior conversations. The gold answer will usually be a concise and short answer that includes the referenced topic, for example:\\{}
Question: Do you remember what I got the last time I went to Hawaii?\\{}
Gold answer: A shell necklace\\{}
The generated answer might be much longer, but you should be generous with your grading - as long as it touches on the same topic as the gold answer, it should be counted as CORRECT.\\{}
\par
For time related questions, the gold answer will be a specific date, month, year, etc. The generated answer might be much longer or use relative time references (like 'last Tuesday' or 'next month'), but you should be generous with your grading - as long as it refers to the same date or time period as the gold answer, it should be counted as CORRECT. Even if the format differs (e.g., 'May 7th' vs '7 May'), consider it CORRECT if it's the same date.\\{}
\par
Now it's time for the real question:\\{}
Question: \$question\\{}
Gold answer: \$golden\_answers\\{}
Generated answer: \$prediction\\{}
\par
First, provide a short (one sentence) explanation of your reasoning, then finish with CORRECT or WRONG. Do NOT include both CORRECT and WRONG in your response, or it will break the evaluation script.\\{}
\par
Just return the label CORRECT or WRONG in a json format with the key as 'label'.
\end{promptbox}

\subsection{Adversarial Scoring Rule}
\label{sec:appx-advrule}

Of LoCoMo's 446 category-5 questions, the \MainbCatFiveNoRef{} that ship no
reference answer are scored correct iff the reply is exactly
\emph{``Not mentioned in the conversation.''}; the \MainbCatFiveAns{} that do ship
one go to the strict judge against the canonical gold. No lexical metric is ever
computed against an adversarial trap answer. The rule was fixed before any
baseline adversarial run (one amended exception in the run
notes: CompassMem's anti-abstention clause was neutralized so the matched arm
is defined).

\end{document}